\documentclass{article}

\usepackage[preprint]{nips_2018}

\usepackage{authblk}
\usepackage{hyperref} 
\usepackage[utf8]{inputenc} 
\usepackage[T1]{fontenc}    
\usepackage{hyperref}       
\usepackage{url}            
\usepackage{booktabs}       
\usepackage{amsfonts}       
\usepackage{nicefrac}       
\usepackage{microtype}      
\usepackage{caption}
\usepackage[labelfont=bf]{caption}
\usepackage{graphicx}
\usepackage{xcolor}
\hypersetup{
    colorlinks,
    linkcolor={blue!50!black},
    citecolor={blue!50!black},
    urlcolor={blue!50!black}
}
\usepackage{todonotes}
\usepackage{comment}
\usepackage{color}
\definecolor{deepmagenta}{rgb}{0.8, 0.0, 0.8}
\definecolor{deepskyblue}{rgb}{0.0, 0.74901960784, 1.0}

\definecolor{red}{rgb}{0.8, 0.0, 0.0}

\definecolor{green}{rgb}{0.0, 0.8, 0.0}

\title{Deep Generative Models in the Real-World:\\An Open Challenge from Medical Imaging}

\newcommand*\samethanks[1][\value{footnote}]{\footnotemark[#1]}
\author[1]{Xiaoran Chen \thanks{Equal contribution}}
\author[2]{Nick Pawlowski \samethanks}

\author[2]{Martin Rajchl}
\author[2]{Ben Glocker}
\author[1]{Ender Konukoglu}
\affil[1]{%
Computer Vision Lab, ETH Zurich}
\affil[2]{%
Imperial College London}

\begin{document}
\maketitle

\begin{abstract}
Recent advances in deep learning led to novel generative modeling techniques that achieve unprecedented quality in generated samples and performance in learning complex distributions in imaging data.  
These new models in medical image computing have important applications that form clinically relevant and very challenging unsupervised learning problems. 
In this paper, we explore the feasibility of using state-of-the-art auto-encoder-based deep generative models, such as variational and adversarial auto-encoders, for one such task: abnormality detection in medical imaging. 
We utilize typical, publicly available datasets with brain scans from healthy subjects and patients with stroke lesions and brain tumors. 
We use the data from healthy subjects to train different auto-encoder based models to learn the distribution of healthy images and detect pathologies as outliers. 
Models that can better learn the data distribution should be able to detect outliers more accurately. 
We evaluate the detection performance of deep generative models and compare them with non-deep learning based approaches to provide a benchmark of the current state of research. 
We conclude that abnormality detection is a challenging task for deep generative models and large room exists for improvement. 
In order to facilitate further research, we aim to provide carefully pre-processed imaging data available to the research community. 

\end{abstract}

\section{Introduction}
\label{intro}
Learning high-dimensional data distributions from finite number of examples and being able to generate new samples from such distributions is a challenging task. 
Developments in deep learning based techniques and unsupervised learning in the last five years set a new standard for this problem, especially for imaging data. 
Generative adversarial networks (GANs)~\cite{goodfellow2014generative}, variational auto-encoders~\cite{kingma2013auto,rezende2014stochastic} and variants of these models~\cite{radford2015unsupervised,arjovsky2017wasserstein,karras2017progressive,makhzani2015adversarial} demonstrate that it is possible to train networks that can approximate distributions of images well enough to sample realistic looking sharp images. 
Such models have already been successfully applied in various vision tasks, such as generating data samples \cite{karras2017progressive}, domain adaptation~\cite{tzeng2017adversarial,chen2018domain} and image in-painting \cite{yeh2016semantic}. 

Unsupervised learning and generative modeling have numerous clinically relevant applications in medical image computing. 
One particular application, \emph{unsupervised abnormality detection}, is scientifically interesting and technically challenging. 
The task is simple to state: given an image acquired from a patient, detect the regions in the image that should not be there in the `normal' case, if any. 
This is one of the routine tasks of a radiologist that they need to perform for every image they assess and a critical first step in diagnosis. 
For complicated cases, years of experience is necessary to distinguish normal from abnormal. 
However, for a large set of problems, such as brain tumors, even non-experts can perform the task after seeing a handful of `normal' looking images. 
Despite the simplicity of its description and the clear separation of abnormal from normal tissue appearance, unsupervised abnormality detection remains as a huge challenge for machine learning. 

Deep learning based generative modeling approaches provide new opportunities for developing automatic algorithms for unsupervised abnormality detection. 
In this work, we empirically investigate feasibility of such approaches using relatively large, publicly available datasets. 
We use magnetic resonance images (MRI) of the brain acquired from healthy individuals at different age groups to train different auto-encoder based generative models to learn the distribution of `normal' brain MRI. 
Then we apply the trained models on two other datasets of brain MRI bearing tumors and stroke lesions to detect the abnormal lesions in an unsupervised manner. 
Detection performance is a good quality indicator that assesses how well the models learn the distribution of `normal' images. 
We describe the datasets and present empirical evaluation comparing different deep learning based models as well as non-deep learning methods, which have been used in the medical image computing community. 
The evaluation provides a benchmark showing the state-of-research for this difficult problem, indicating that unsupervised detection of abnormalities remains an open challenge and demands further research.

\section{Related work}
\noindent\textbf{Related work on deep generative models:} Literature on generative modeling with neural networks date back to MacKay's work \cite{mackay1995bayesian}. More recent GANs \cite{goodfellow2014generative} and VAEs \cite{kingma2013auto,rezende2014stochastic} shown the feasibility of generative modeling with deep models. Further modifications suggested in \cite{radford2015unsupervised} \cite{arjovsky2017wasserstein} \cite{gulrajani2017improved} improved the performance and stability of GAN training, and achieve realistic data generation such as \cite{karras2017progressive}. Alternatively, VAEs mostly rely on a reconstruction loss that is known to lead to blurry reconstructions \cite{larsen2015autoencoding}. In contrast to GANs, VAE-based works mainly focus on the latent variable model \cite{gregor2015draw,yan2016attribute2image} and how to disentangle the latent variables \cite{chen2018isolating}. Many publications propose methods to improve the expressibility of this model \cite{rezende2015variational,kingma2016improved,burda2015importance}.

\noindent\textbf{Lesion detection and segmentation:}
Detection of brain lesions is a critical step to diagnose diseases such as cranial trauma, abscesses and cancer. Traditionally, radiologists manually detect and segment lesions slice by slice. However, the large resolution of the 3D images and high level of required expertise have made it a time-consuming and expensive task to accomplish. Studies such as \cite{prastawa2004brain}, \cite{ayachi2009brain}, and \cite{zikic2012context} have suggested supervised methods for automatic detection of brain lesions. Due to the importance of the application, the medical image computing community has been hosting public challenges specifically for lesion detection, the Multi-modal Brain Tumor Image Segmentation (BRATS) and Ischemic Stroke Lesion Segmentation (ISLES). A benchmark \cite{bauer2012segmentation} was released to evaluate and compare existing models. With the introduction of fully convolutional neural networks (FCNs) \cite{long2015fully}, DeepMedic \cite{kamnitsas2017efficient} and U-Nets \cite{ronneberger2015u}, the field has since then been dominated by deep learning approaches achieving the highest accuracy on most if not all imaging challenges.

Supervised methods however, are specific to certain lesions and need to be trained with respective example data. Furthermore, their application on previously unseen lesions is not straightforward. 

\noindent\textbf{On unsupervised lesion detection: }
Unsupervised detection of abnormalities has been an important topic in medical image computing. Non-deep learning based methods have been proposed over the last two decades that use mixture modeling and expectation maximization~\cite{van2001automated}, atlas-registration~\cite{prastawa2004brain} and probabilistic models that utilize image registration~\cite{tomas2015model,zeng2016abnormality}. 

Deep-learning based models have also been recently applied to abnormality detection following related developments in computer vision\cite{an2015variational, chalapathy2017robust}. Schlegl et al. used GANs to detect abnormalities in \cite{schlegl2017unsupervised}. Their method was based on determining the best latent space representation of a given image with abnormality and then computing the difference between reconstruction from this representation and the image. Underlying idea was that GAN trained on healthy images should not be able to reconstruct abnormal lesions. Based on related ideas, more recent work explored the use of auto-encoder based models and detecting abnormal regions through reconstruction error \cite{sato2018primitive,baur2018deep,chen2018unsupervised,pawlowski2018unsupervised}.

\section{Methodology}
\label{methods}
We approach the lesion detection problem in a way similar to one-class classification, where firstly we model the pixel-wise probability using healthy brain MRI images, then detect lesion regions as pixels with low probability according to the model learned on healthy data. Assume we have a dataset of healthy images $\{X_{H}^{1}, \dots, x_{H}^{N}\}$ where each image is a set of pixels $X_H^{n} \in R^{d\times d} = \{x_H^{(n, 1)}, \dots, x_H^{(n, P)}\}$. Given this dataset, we aim to estimate the distribution of healthy data $P_H$ to evaluate the probability of an unseen image and its pixels $P_H(x^{p})$. 
Now, suppose we have another test image $X_A \in R^{d\times d}$ with both abnormal $p_{A}$ and healthy regions $p_{H}$. Because $P_H$ only models the distribution of healthy image, the probability of the pixels in abnormal regions $P_H(x_A^{(p \in p_A)})$ are expected to be low, whereas pixels in healthy regions will have high probability $P_H(x_A^{(p \in p_H)})$. 

A naive approach could be to model $P_H$ as a location dependent function with a Gaussian per pixel $P_H(\hat{x})=P_H(\hat{x}\mid i, j)=\mathcal{N}(\hat{x}\mid\mu_{(i, j)}, \sigma_{(i, j)})$, where $i$ and $j$ denote the location of the pixel in the image and $(\mu_{(i, j)}, \sigma_{(i, j)})$ are the parameters of the Gaussian corresponding the pixel $(i, j)$. By assuming $\sigma_{(i, j)} = 1$ this becomes the difference of a new pixel $\hat{x}$ to its corresponding mean pixel value from the healthy images. 

A more advanced approach would be to model pixel intensities at each location with a Gaussian Mixture Model (GMM) where probability of an unseen pixel is modeled as $P_H(\hat{x})$ as  $P_H(\hat{x}) = \sum\limits_{i=1}^{K} \Phi^j_i \mathcal{N}(\hat{x}\mid\mu_i, \sigma_i)$ with the number of mixture components $K$, the mixture weights $\{\Phi^j_i\}$ that depends on location $j$, and the parameters of each mixture component $(\mu_i, \sigma_i)$. Given an image, the parameters of this model can be estimated specific to image using Expectation-Maximization algorithm and using an atlas image. Furthermore, abnormality detection can be performed by adding an additional component whose $\Phi^j_o P(\hat{x} | o) = \lambda$ is a constant for all pixels and intensities similar to what is proposed in~\cite{van2001automated}. 


AE-based models consist of two deterministic mappings, the encoder $f_{enc}$ and the decoder $f_{dec}$. An input $\hat{x}$ goes through $f_{enc}$ to be encoded into a lower dimensional latent variable $z$, and then goes through $f_{dec}$ to be decoded back into an reconstruction $\hat{X}' = AE(\hat{X})$ that is based on the latent encoding $z$. The functions $f_{enc}$ and $f_{dec}$ are then optimized to minimize the reconstruction loss $L(\hat{X}, \hat{X}')$. We chose to employ the frequently used $L_2$ loss as reconstruction loss $L(\hat{X}, \hat{X}') = ||\hat{X}-\hat{X}'||_{2}$. Due to the lower dimensionality of $z$, AE-based methods are forced to learn a compression of the data that is related to learning a lower manifold representation. 

We argue that because the AE relies on this lower dimensional representation it is not capable of reconstructing variations in the data that it has not seen during training. Therefore the reconstruction loss $L(\hat{X}, \hat{X}')$ can be interpreted as an unnormalized probability of the a sample belonging to the data distribution $P_H(\hat{X}) = 1/Z L(\hat{X}, AE(\hat{X}))$ where $Z$ is an unknown normalization constant.

Various adaptations of the basic AE have been suggested. De-noising AE follow the same concept as AEs but aim to reconstruct clean images $\hat{X}$ from corrupted images $\hat{X}+\sigma$. By trying to remove noise from the images, a DAE distinguishes noise from structure in $\hat{X}$, thus better captures the information in them. However, AE and DAE are not generative models as they do not approximate $P_H$ and merely serve as dimensionality reduction methods. Variational AEs (VAEs) and Adversarial AEs (AAEs) integrate stochastic inference into the AE framework and enable to approximately model $P_H$ via variational inference. The deterministic mappings $f_{enc}$ and $f_{dec}$ in regular AEs, become probabilistic mappings and model the inference network $Q(z\mid X)$ and generative process $P(X\mid z)$ respectively. The distribution is learned in such a way that, 
\begin{equation}
P_H(X_H) = \int P(X_H\mid z_H)P(z_H) dz,
\end{equation}
where $P(z_H)$ describes a prior on the latent encoding that constrains $z_H$ to lie in a structured latent space. The structured latent space is imposed on $Q(z\mid X)$ by minimizing a Kullback-Leibler (KL) divergence $KL[Q(z|X)||P(z)]$. The overall model is optimised by maximizing the evidence lower bound (ELBO) \cite{doersch2016tutorial},
\begin{equation}
\log P_H(X_H) \geq E_{z\sim Q(z_H\mid X_H)}[\log P(X_H\mid z_H)] - KL[Q(z_H\mid X_H)||P(z)]
\end{equation}

Several studies found VAEs to generate blurry reconstructions \cite{larsen2015autoencoding}\cite{bousquet2017optimal}. As good quality reconstructions is potentially a prerequisite for satisfactory detection outcomes, we seek alternatives to mitigate the blurriness. \cite{bousquet2017optimal} suggests that AAE resolve the blurriness by improving the encoder using adversarial learning to match $Q(z) = E_{X_H}[Q(z|X_H)]$ and $P(z)$ by optimizing a GAN loss, 
\begin{equation}
\label{eq3}
\min_G \max_D E_{z\sim P(z)}[\log D(z)] + E_{z\sim Q(z)} [\log(1-D(z))]
\end{equation}
where the generator $G$ corresponds to the encoder in AAE. 
To stabilized GAN training, we modify the loss of the original AAE to use the recently proposed Wasserstein distance from WGANs with gradient penalty (WGAN-GP) \cite{gulrajani2017improved}.

Another approach to achieve sharper images with AE-based methods is to improve how the image is compared to its reconstruction. In the work of $\alpha$-GAN, the model matches $Q(z)$ and $P(z)$ in the same way as AAE and adds one more discriminator $D_{rec}$ to distinguish between $X_H$ and $X'_H$. Again, the addition of $D_{rec}$ introduces an adversarial loss that can be written in a similar form as Eq. \ref{eq3}. Here, the decoder acts as the generator in the GAN formulation. The optimization is more complicated due to the modifications above. To train the model, we follow the optimization provided in \cite{rosca2017variational}.

Lastly, we employ variational inference to approximate the posterior distributions over the model parameters with factorized Gaussians \cite{blundell2015weight}. This allows us to not only marginalize the latent encoding but also the model parameters when estimating the reconstruction loss of a new image. This might enable more robust reconstructions as it is less reliant on specific model parameters as it has been shown by \cite{pawlowski2018unsupervised}.

\section{Experiments}
To give a comprehensive overview of the current state of unsupervised lesion detection, we also include baselines like GMMs and mean image difference and further test a supervised segmentation task using an U-Net \cite{ronneberger2015u}. 

\subsection*{Data}
\label{data}
\textbf{Cam-CAN}\footnote{\url{http://www.cam-can.org/}}\cite{taylor2017cambridge}
We use The Cambridge Centre for Ageing and Neuroscience (Cam-CAN) dataset for training which contains T1- and T2-weighted brain MRI of 652 subjects from a uniform age range from 18--87. All subjects are confirmed healthy after radiological assessment.

\textbf{BraTS-T2w}\footnote{\url{https://www.med.upenn.edu/sbia/brats2018.html}}\cite{menze2015multimodal,bakas2017advancing}
We utilize the T2-weighted images of 285 patients from the Brain Tumor Segmentation Challenge (BraTS). The images show high-grade (210) and low-grade (75) glioblastomas which are visible as brighter regions in the images.

\textbf{ATLAS-T1w}\footnote{\url{http://fcon_1000.projects.nitrc.org/indi/retro/atlas.html}}\cite{liew2018large}
We also make use of the Anatomical Tracings of Lesions After Stroke (ATLAS) dataset containing T1-weighted images of 220 stroke patients. Lesions are visible as darker regions in the images and identified using location information as appearance is similar to normal structures.

\begin{figure}[!ht]
\centering
  \includegraphics[scale=0.3]{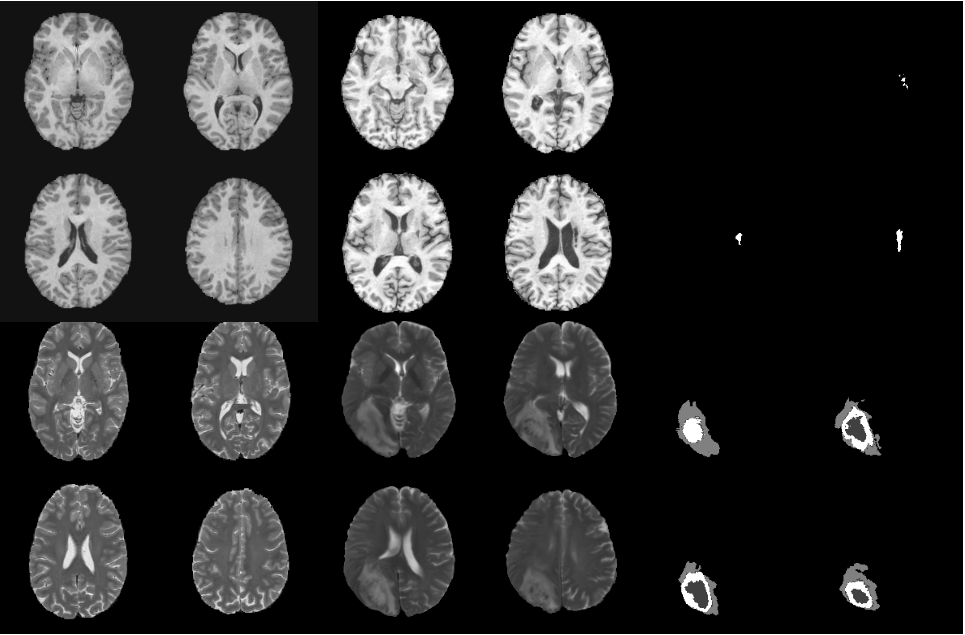}%
  \caption{Thumbnails of datasets. From left to the right, Column 1-2 are CamCAN dataset with upper two rows being T1-weighted images and bottom two rows being T2-weighted images, column 3-4 present testing datasets with upper two rows being images from ATLAS-T1w datasets and bottom two rows being images from BraTS-T2w dataset, column 5-6 are ground truth segmentations for the corresponding images in column 3 and 4.}
\end{figure}

\subsection*{Pre-processing}
To reduce the variability across subjects and datasets, each scan is normalized as follows. First, empty images with no brain are removed, and then, the images are cropped within the maximal boundary computed across the dataset to ensure the same image size, lastly, the images are normalized to have zero-mean and unit-variance within the brain masks obtained from skull-stripping process. The models are trained with datasets of two different sizes, $128 \times 128$ and $256 \times 256$. Resizing is implemented using \verb|scipy.misc.imresize| with nearest interpolation in python package.

\subsection*{Evaluation}
For all difference methods we calculate the difference of a new image $\hat{X}$ and its reconstruction $\hat{X}'$ as the absolute error $\hat{X}_{dif} = |\hat{X}-\hat{X}'|$ instead of the squared error. This does not change the outcome of the predictions as both are monotonic functions and we use thresholding to find abnormal regions. We evaluate $\hat{X}_{dif}$ using the ground truth annotations of lesions. Let the ground truth be $Y$. We use the following metrics for detection performance evaluation of the trained models,

\begin{enumerate}
\item \textbf{Area Under Curve (AUC)}. The AUC is calculated as the area under ROC curve due its insensitivity to label imbalance that occurs in our dataset. In particular, we compute the true positive rate (TPR) $TPR=\frac{TP}{TP+FN}$ and false positive rate (FPR) $FPR=\frac{FP}{FP+TN}$.
\item \textbf{Maximal Dice Score (mDSC)}. The dice score is commonly used to report segmentation results. To calculate the dice score in our case, it is required to set a threshold $t$ for $\hat{X}_{dif}$ that predicts lesions as $\hat{Y} = \hat{X}_{dif} > t$. While it can be formed into a new question, we use a range of thresholds and calculate a dice score using $\hat{X}_{dif}$ and $Y$ for each threshold. As we do not further explore thresholding, we assume there is an optimal threshold that achieves the maximal dice score (mDSC) on the model. In other words, mDSC depicts the maximal achievable dice score of a model. This also measures the separability of the distributions of the reconstruction error of healthy and abnormal tissue.
\end{enumerate}
\section{Results}
\label{results}
\begin{figure}[!ht]
\captionsetup{width= 130mm}
\begin{minipage}{.5\textwidth}
\centering
\includegraphics[scale=0.2]{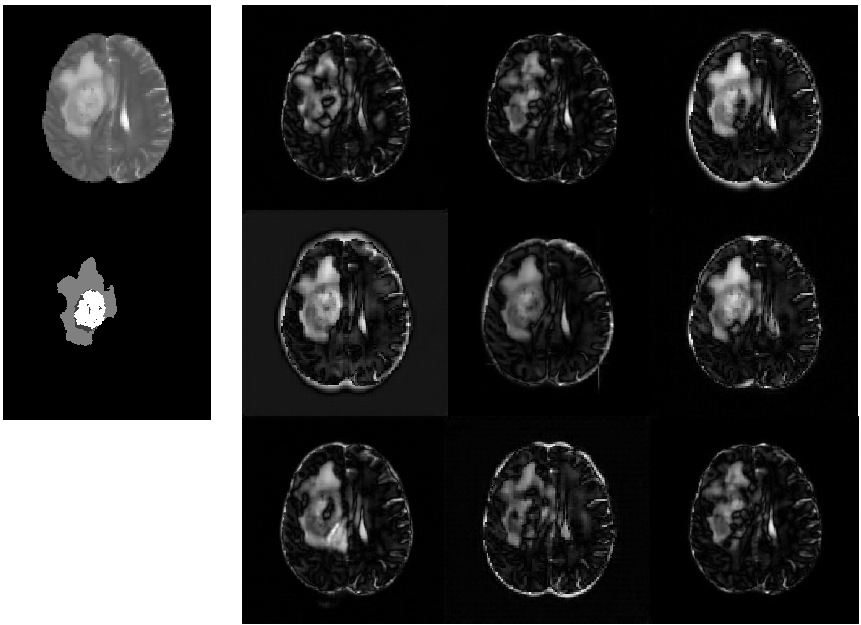}
\end{minipage}
\begin{minipage}{.5\textwidth}
\centering
\includegraphics[scale=0.5]{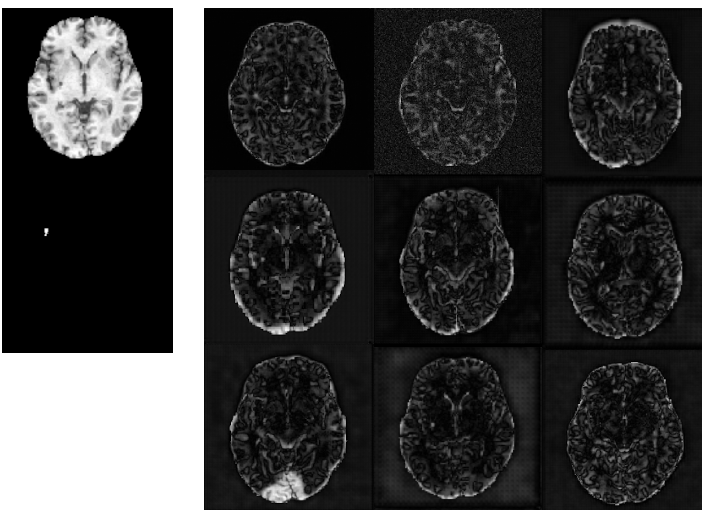}
\end{minipage}
\caption{Difference map obtained on BraTS-T2w dataset (left) and ATLAS-T1w (right). For each dataset, the leftmost column shows the original image (upper) and ground truth segmentation (lower). Columns 2-4 show difference obtained with auto-encoding models as follows, row 1: AE, DAE, VAE-128, row 2: VAE-BBB-128, VAE-256, AAE-128, row 3: AAE-256, $\alpha$-GAN-128, $\alpha$-GAN-256}
\label{fig:dif_map}
\end{figure}

We demonstrate our results by firstly presenting the difference maps as in Figure \ref{fig:dif_map} for visual inspection of the models and then providing quantitative results using the metric mentioned in section \ref{methods} for detailed comparison. Here, we show difference map to represent the reconstruction error $X_{dif}$. Due to a known domain gap, models are trained respectively for the two available modalities, T1-weighted and T2-weighted on healthy brain images in CamCAN datasets. Trained models are later tested on its corresponding modality on BraTS-T2w and ATLAS-T1w. Thus in the work, we consider two independent detection tasks, 1) detection of tumor on BraTS-T2w, 2) detection of lesion on ATLAS-T1w. All models are trained until convergence on GPU cluster with GeForce GTX TITAN X (12207 MiB). To calculate the metric of mDSC, we iterate the conventional dice score calculation through an arbitrary range [0.0,6.0] with 1001 interval.  As GMM models output probability maps with values between 0.0 and 1.0, the range is changed to [0.0,1.0] with 400 intervals. Note that mDSC is more of a numerical approximation of the maximal dice score by brutal-force search. The value does get better with more intervals whereas optimization may be needed to efficiently approximate the optimal value.

For the convolution version of VAE and AAE, latent variables are obtained from the previous convolution layer instead of a dense layer to avoid possible loss of spatial information. Following the theory of vanilla VAE, we still assume each latent variable is independent and compute KL divergence between $Q(z_H|X_H)$ and $\mathcal{N}(0,I)$. 

As auto-encoder-based methods usually have difficulty reconstructing large images, such as images of size 256 $\times$ 256, we reduce the challenge by training our models also on down-sampled datasets. The down-sampled datasets are obtained for both training and test datasets by resizing the original images to the size of 128$\times$128 as described in section \ref{data}. However, experiment results achieved by training and testing on down-sampled datasets are not significant better but rather similar to that on the original datasets. Although down-sampling has little advantage in terms of performance, it has more impact on training as it requires shorter running time and less memory compared to the original datasets. 

As the tumor shows relatively high intensity on T2-weighted images, the resulting intensity difference between the healthy and abnormal can be more obvious than the intensity difference of lesions viewed in T1-weighted images. Additionally, the size of tumor in BraTS is often larger than that of the lesion in ATLAS. This property may, to some extent, ease the difficulty of tumor detection with BraTS-T2w. In the difference maps obtained on BraTS-T2w and ATLAS-T1w, we can confirm this statement. Figure \ref{fig:dif_map} shows that the tumor can be often fully or partially highlighted by the auto-encoding methods while the detection appears worse on ATLAS-T1w, indicating a more impressive detection outcome on BraTS-T2w.



Our baseline methods appear to be strong baselines with good performance. GMM achieves the highest AUC on both datasets, although its mDSC on BraTS-T2w appears less satisfactory. The superior performance of GMM model might be due to the fact that it is not entirely unsupervised as the number of components are pre-defined based on anatomical knowledge. U-net, as a widely-favored supervised method, achieves the highest dice score on both datasets. Comparison among the models shows similar performances on BraTS-T2w, and similarly on ATLAS-T1w. The results are consistent with the difference maps. Detection on BraTS-T2w indicates that auto-encoder-based models are capable of detecting the large-size abnormality although mDSC can be further improved. In contrast, the results on ATLAS-T1w imply difficulty of detection using either unsupervised or supervised methods. In terms of models, although none of them has a significant advantage over the rest on both tasks, AAE, VAE and VAE-BBB are the most effective ones with higher accuracy than the others. From here, we describe the detection results for each dataset. 

\begin{table}[h!]
  \caption{Summary of evaluation metrics with AUC and mDSC with parameters settings. Specifically, we train DAE with Gaussian noise  $\mathcal{N}(\mu=0,\sigma=0.5)$. VAE-128 and VAE-256 denote VAE trained on datasets in the image size of 128$\times$128 and 256$\times$256 respectively, and likewise for AAE and $\alpha$-GAN. Baseline of GMM is provided under two parameter settings, $\lambda_{out}$=0.01 and $\lambda_{out}$=0.001. Latent variables are shown as tensors for convolutions auto-encoding models.}
  \label{metric table}
  \centering
  \begin{tabular}{lc|cc|cc}
    \toprule
    & Latent   & \multicolumn{2}{c}{BraTS-T2w}  & \multicolumn{2}{c}{ATLAS-T1w}                  \\
    & variables &\multicolumn{2}{c}{(whole tumor)}  & &\\
     \cmidrule(r){2-6}  
    Models        &z   & AUC    & mDSC  & AUC  & mDSC\\
    \midrule
mean          & -  &   0,65    & 0.20   &  0.46 & 0.02 \\
AE           & 256  &  0.63    & 0.41   &  0.49 & 0.03 \\
DAE ($\sigma$=0.5)   &256   & 0.59   &  0.29 & 0.41 & 0.06  \\
VAE-128    & (2,2,64)  & 0.69   &  0.42 & 0.64 & 0.08  \\
VAE-BBB-128&(2,2,64)   & 0.69   &  0.40 & 0.67 & 0.05 \\  
VAE-256      & (4,4,64)  & 0.67   &  0.40 & 0.66 & 0.08  \\
AAE-128      &(2,2,64)  & 0.70   & 0.41   & 0.63 & 0.06  \\
AAE-256      &(4,4,64)  & 0.67   &  0.38 & 0.60 & 0.04 \\
$\alpha$-GAN-128 & 128    & 0.66   &  0.35 & 0.60 & 0.05 \\
$\alpha$-GAN-256 & 256    & 0.67   &  0.37 & 0.60 & 0.04\\
\midrule
GMM ($\lambda_{out}$=0.01)  & -  & 0.80  &  0.22  & 0.78 & 0.17 \\
GMM ($\lambda_{out}$=0.001) & -  & 0.79  &  0.21  & 0.77 & 0.17\\
    U-net (supervised) & - &    -     & 0.80 &  -    & 0.50  \\
    \bottomrule
  \end{tabular}
\end{table}

\begin{figure}[!ht]
\captionsetup{width= 130mm}
\begin{minipage}{.5\textwidth}
\centering
\includegraphics[width=1.\textwidth]
{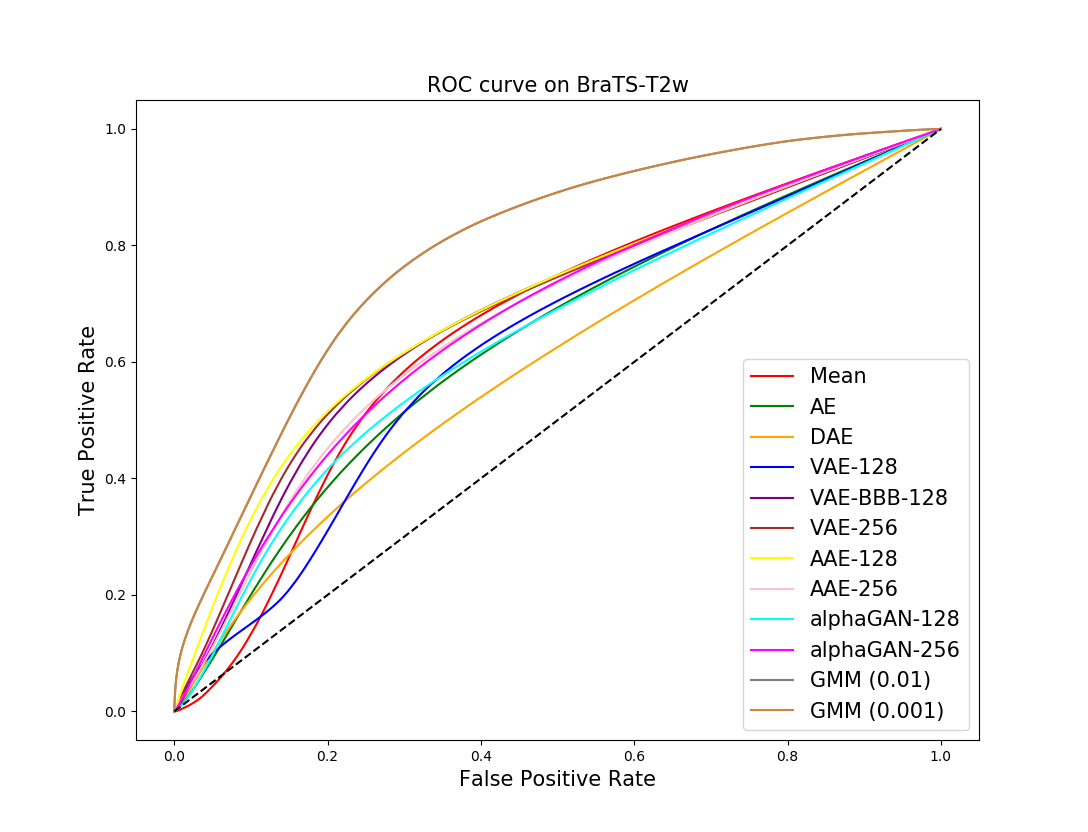}
\end{minipage}
\begin{minipage}{.5\textwidth}
\centering
\includegraphics[width=1.\textwidth]{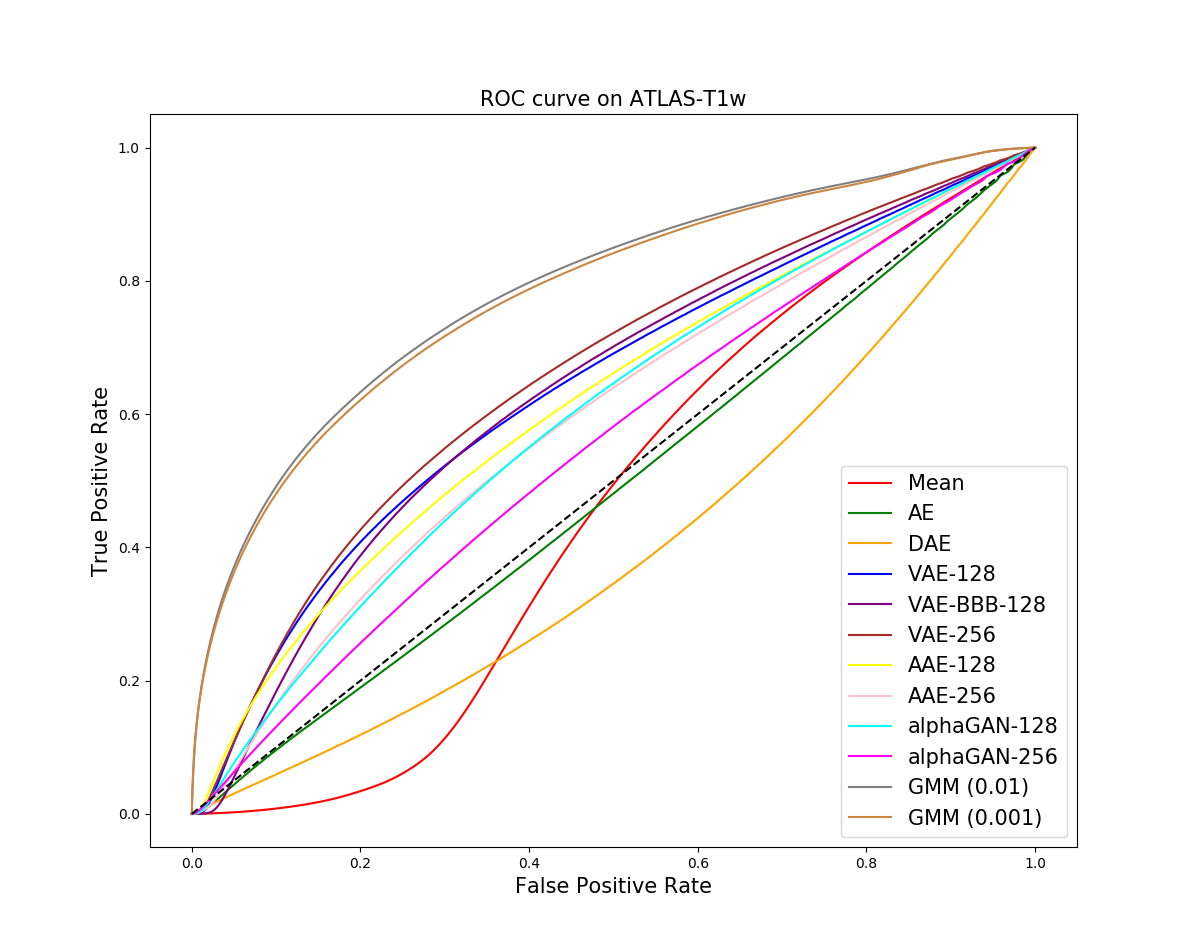}
\end{minipage}
\label{fig:roc_curve}
\caption{ROC curves corresponding to reported AUC in Table \ref{metric table}. On the right is ROC curves for models trained on BraTS-T2w, and on the left is the plot for models trained on ATLAS-T1w. Black dashes line marks the performance of random classification. Effective models should curves above this dashed line. Note that GMM(0.01) and GMM(0.001) have almost indistinguishable performance on BraTS-T2w which causes their curves to overlap.}
\end{figure}


\subsection*{BraTS-T2w}
In terms of AUC score, convolutional VAE and VAE-BBB trained on down-sampled 128 $\times$ 128 datasets have the highest value. Convolution VAE outperforms VAE-BBB on mDSC marginally by 2\% achieving the highest score in this metric. The second highest is achieve by convolutional AAE trained on down-sampled dataset with 0.41, which is 1\% lower than convolution VAE. DAE has significantly inferior performance compared the most effective three. Although $\alpha$-GAN theoretically produces realistic and sharp images, the models trained on down-sampled and original datasets do not show advantage in this detection task. $\alpha$-GAN have similar but slight lower AUC score than the best-performing models while its mDSC is lower than the highest by 0.5 and 0.7 respectively on down-sampled and original datasets.



\subsection*{ATLAS-T1w}
The auto-encoder-based models achieved worse performance on ATLAS-T1w than on BraTS-T2w. Although AUC metric, which is calculate with TPR and FPR, does not show significant decrease, mDSC reveals the weakness of the models on detecting lesion in T1-weighted images. None of the models achieves comparable results as in the detection tasks with T2-weighted images. The mDSC achieved by each model is below 0.1, indicating that the lesion cannot be distinguished from the normal structures by any threshold we have applied to $X_{dif}$ in the experiments. In comparison, dice score of U-net and mDSC of GMM are significantly higher than that of the auto-encoder based methods. In spite of this large performance gap, supervised segmentation with U-net only achieves dice score of 0.50, which is relatively low for a supervised segmentation task. Given the results on ALTAS-T1w, we conclude that the unsupervised detection of lesion on ATLAS-T1w remains a hard task where supervised segmentation is also difficult. 

\section{Discussion}
In this work, we provide an overview of the current state of unsupervised outlier detection on standard medical images. We evaluate auto-encoder-based unsupervised models in terms of AUC and mDSC to describe their strength on this new application. Results indicate that convolutional VAE, Bayesian VAE and AAE have great potential to be further studied and developed to gain higher detection accuracy. We also identify that detection with T2-weighted datasets may be an easier first step to explore modification and seek improvement while detection with T1-weighted remains more challenging in both unsupervised and supervised ways. Besides selection of models, we observe in our experiments that reconstruction with data cropping as described in \ref{data} produces lower reconstruction error than without cropping. This simple preprocessing step reduces the area of uninformative background and prevents the models from taking the shortcut of reducing reconstruction error by learning to reconstruct background. Moreover, it is easily noticed that the performance achieved by current available auto-encoding models is worse than the popular supervised methods U-net. We suggest some possible directions to bring improvements.

\textbf{Improvement in reconstruction quality} As the detection is based on absolute reconstruction error, it is straight-forward that higher accuracy can be achieved if the model is able to obtain sharp and accurate reconstruction. One of many approaches to achieve this is by combination of VAE and GAN to produce sharp images as several works have suggested. 

\textbf{Estimation of pixel-wise probability} In our approach, the pixel-wise probability is approximated by calculating reconstruction error. As is seen in the difference map, the models manage to reconstruct an image with abnormality as a healthy-looking image, which is within our expectation for the models. Although the reconstruction appears to be healthy-looking, taking absolute intensity difference may be rigid because this difference ignores structural differences. Assume there exists such a pixel $X^a_A$ which is an abnormal pixel and the pixel $X^h_A$ which is a high-intensity normal pixel, the case where $X^a_A-X'^a_A \leq X^h_A-X'^h_A$ can occur even if $X'^a_A$ is reconstructed into a normal pixel. When such cases are prevalent in the dataset, this largely lowers the performance even if the reconstruction is of good quality. This is to say, a more proper pixel-wise probability estimation can be helpful to improve performance.

\textbf{Thresholding} Another question lies in selection of threshold. In this work, we leave the selection of thresholds open and instead evaluate the models within a range of thresholds. One may argue that Dice score can be calculated using statistical measure such as 90\% percentile. Thresholding according to a given percentile can be valid whereas the percentile may not be optimal for the data. Selection of a proper and adaptive threshold can also help to distinguish outliers from normal structure.

\small
\bibliographystyle{unsrt}
\bibliography{nips_2018}
\end{document}